\documentclass[final]{cvpr}

\usepackage{times}
\usepackage{epsfig}
\usepackage{graphicx}
\usepackage{amsmath}
\usepackage{amssymb}
\usepackage{multirow}
\usepackage{booktabs}
\usepackage{pifont}

\newcommand{\cmark}{\ding{51}}
\newcommand{\xmark}{\ding{55}}

\usepackage[pagebackref=true,breaklinks=true,colorlinks,bookmarks=false]{hyperref}

\def\confYear{CVPR 2021}

\begin{document}

\title{APES: Audiovisual Person Search in Untrimmed Video}

\author{ \normalsize Juan León Alcázar$^{1}$, Long Mai$^{2}$, Federico Perazzi$^{2}$, \\ \normalsize Joon-Young Lee$^{2}$, Pablo Arbeláez$^{3}$, Bernard Ghanem$^{1}$, and Fabian Caba Heilbron $^{2}$  \\ 
\normalsize $^{1}$Universidad de los Andes, $^{2}$Adobe Research, $^{3}$King Abdullah University of Science and Technology, \\
\scriptsize $^{1}$\{\texttt{juancarlo.alcazar,bernard.ghanem}\}\texttt{@kaust.edu.sa}
\scriptsize $^{2}$\{\texttt{caba,malong,perazzi,jolee}\}\texttt{@adobe.com}
\scriptsize $^{3}$\{\texttt{pa.arbelaez}\}\texttt{@uniandes.edu.co} 
\\ }

\maketitle

\begin{abstract}
\vspace{-0.25cm}
    Humans are arguably one of the most important subjects in video streams, many real-world applications such as video summarization or video editing workflows often require the automatic search and retrieval of a person of interest. Despite tremendous efforts in the person re-identification and retrieval domains, few works have developed audiovisual search strategies. In this paper, we present the Audiovisual Person Search dataset (APES), a new dataset composed of untrimmed videos whose audio (voices) and visual (faces) streams are densely annotated. APES contains over 1.9K identities labeled along 36 hours of video, making it the largest dataset available for untrimmed audiovisual person search. A key property of APES is that it includes dense temporal annotations that link faces to speech segments of the same identity. To showcase the potential of our new dataset, we propose an audiovisual baseline and benchmark for person retrieval. Our study shows that modeling audiovisual cues benefits the recognition of people's identities. To enable reproducibility and promote future research, the dataset annotations and baseline code are available at: \href{https://github.com/fuankarion/audiovisual-person-search}{ https://github.com/fuankarion/audiovisual-person-search}
\end{abstract}

\section{Introduction}
\label{sec:intro}

Can we find every moment when our favorite actor appears or talks in a movie? Humans can do such search relying on a high-level understanding of the actor's facial appearance while also analyzing their voice \cite{bruce1986understanding}. The computer vision community has embraced this problem primarily from a visual perspective by advancing face identification \cite{facenet, parkhi2015deep, tapaswi2019video}. However, the ability to search for people using audiovisual patterns remains limited. In this work, we address the lack of large-scale \textit{audiovisual} datasets to benchmark the video person retrieval task. Beyond finding actors, several video domain applications could benefit from our dataset, from accelerating the creation of highlight moments to summarizing arbitrary video data via speaker diarization.

Contrary to image collections, video data casts additional challenges for face and person retrieval tasks \cite{grother2017face}. Such challenges include drastic changes in appearance, facial expressions, pose, or illumination as a video progresses. These challenges have fostered research in video person search. Some works have focused on person re-identification in surveillance videos \cite{gheissari2006person, farenzena2010person,market}. In this setup, the goal is to track a person among a set of videos recorded from various cameras, where the global appearance (\eg clothing) of the target person remains constant. Another setup is the cast search problem, where models take a portrait image as a query to retrieve all person tracks that match the query's identity \cite{csm}. The community has achieved relevant progress, but the lack of large scale audiovisual information still prevents the development of richer multi-modal search models.

\begin{figure*}[t!]
    \begin{center}
        \includegraphics[width=0.99\textwidth]{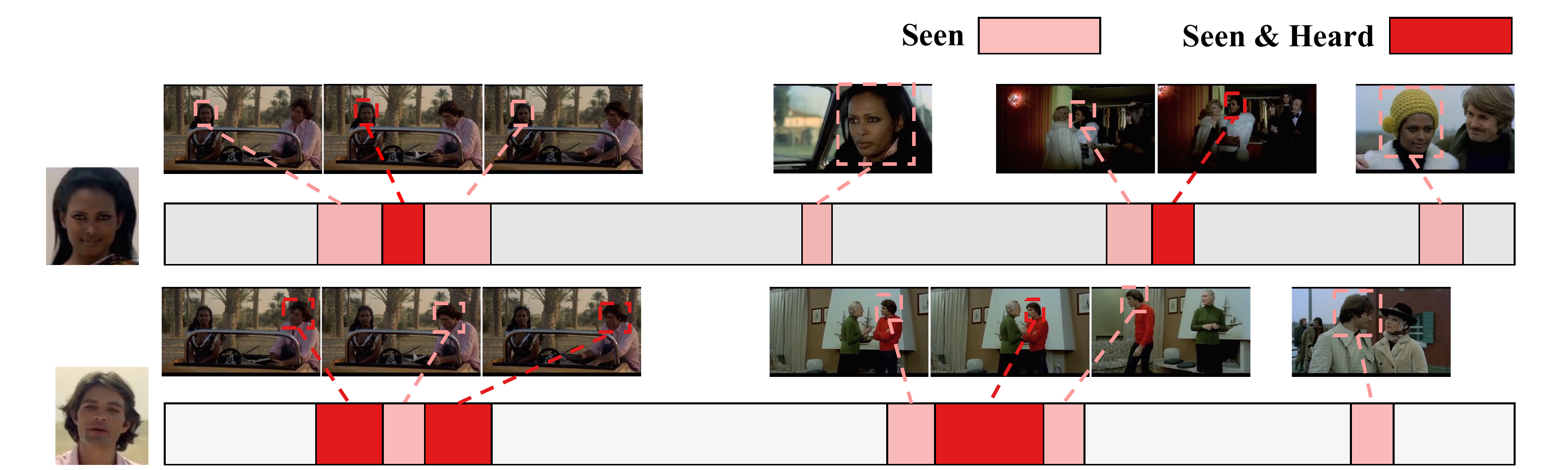}
    \end{center}
    \caption{\textbf{Audiovisual PErson Search (APES).} We introduce APES, a novel video dataset for multimodal person retrieval. The dataset includes annotations of identities for faces and voice segments extracted from untrimmed videos. We establish three new person search tasks. (i) Seen aims to retrieve all timestamps when a target person is on-screen.  (ii) Seen \& Heard, which focuses on finding all instances when a person is visible and talks.}
    \label{fig:intro}
\end{figure*}

Motivated by PIN cognitive models \cite{bruce1986understanding}, Nagrani \etal \cite{learnablepins} have developed self-supervised models that learn joint face-voice embeddings. Their key idea is to use supervision across modalities to learn representations where individual faces match their corresponding voices via contrastive learning \cite{contrastive}. However, many videos in the wild might contain multiple visible individuals that remain, mostly, silent. This situation introduces a significant amount of noise to the supervision signal \cite{ava-active-speakers}. Liu \etal \cite{iqiyi} have also explored audiovisual information for person retrieval in videos. To this end, their work introduces the iQIYI-VID dataset, which contains videos from social media platforms depicting, in large proportion, Asian celebrities. Despite its large-scale, the dataset contains only short clips, with most of them being five seconds long or shorter. We argue that having long videos is crucial to high-level reasoning of context to model real-life expressions of people's faces and voices. Additionally, we require densely annotated ground-truth labels to enable direct links between speech and visual identities.

In this paper, we introduce APES (Audiovisual PErson Search), a novel dataset, and benchmark for audiovisual person retrieval in long untrimmed videos. Our work aims to mitigate existing limitations from two angles. First, we densely annotate untrimmed videos with person identities and match those identities to faces and voices. Second, we establish audiovisual baselines and benchmarks to facilitate future research with the new dataset. Our dataset includes a broad set of 15-minutes videos from movies labeled among a long-tailed distribution of identities. The dataset samples account for many challenging re-identification scenarios such as small faces, poor illumination, or short speech segments. In terms of baselines, we develop a two-stream model that predicts people's identities using audiovisual cues. We include benchmarks for two alternative tasks. \textit{Seen}, which aims at retrieving all segments when a query face appears on-screen; \textit{Seen \& Heard}, which focuses on finding instances where the target person is on-screen and talking. Figure \ref{fig:intro} showcases APES annotations and tasks.

\paragraph{Contributions.} This paper's primary goal is to push the envelope of audiovisual person retrieval by introducing a new video dataset annotated with face and voice identities. To this end, our work brings two contributions.
\begin{itemize}
    \item We annotate a dataset of $144$ untrimmed videos from movies. We associate more than $30K$ face tracks with $26K$ voice segments and label about $1.9K$ identities. Section \ref{sec:dataset} details our data collection procedure and showcases the characteristics of our novel dataset.
    \item We establish an audiovisual baseline for person search in Section \ref{sec:study}. Our study showcases the benefits of modeling audiovisual information jointly for video person retrieval in the wild.
\end{itemize}
\section{Related Work}

\label{sec-related}
There is a large corpus of related work on face \cite{tapaswi2019video, facenet, parkhi2015deep} and voice \cite{zhang2019fully, voxceleb2} retrieval. This section focuses on related work on datasets for video person retrieval and audiovisual learning for identity retrieval.

\begin{table*}[t!]
    \setlength{\tabcolsep}{3pt}
    \centering
    \begin{tabular}{ l c c c c c }
    \toprule
    Dataset & Source & Task & $\#$ Instances & $\#$ Tracklets & $\#$ Identities \\
    \midrule
    \midrule
    PSD \cite{psd} & Images & Re-identification & 96K & - & 8432 \\
    Market \cite{market} & Images & Re-identification & 32K & - & 1501 \\
    MARS \cite{mars} & Person Tracks & Re-identification & 1M & 20K & 1261 \\
    \midrule
    VoxCeleb \cite{voxceleb} & Short Clips & Speaker Recognition & - & 22.5K & 1251 \\
    VoxCeleb2 \cite{voxceleb2} & Short Clips & Speaker Recognition & - & 150K & 6112 \\
    \midrule
    iQIYI-VID \cite{iqiyi} & Short Clips & Visual Search & 70M & 600K & 5000\\
    CSM \cite{csm} & Person Tracks & Visual Search & 11M & 127K & 1218 \\
    \midrule
    Big Bang Theory \cite{bigban} & Untrimmed Videos & Audiovisual Search & - & 3.7K & 8 \\
   Buffy \cite{everingham2006hello} & Untrimmed Videos & Audiovisual Search & 49k & 1k & 12 \\
    Sherlock \cite{sherlock} & Untrimmed Videos & Audiovisual Search & - & 6.5K & 33 \\
    \textbf{APES} & Untrimmed Videos & Audiovisual Search & 3.1M & 30.8K & 1913  \\ 
    \toprule
    \end{tabular}
    \caption{
        \textbf{Video person identity retrieval datasets overview.} 
        APES is the largest dataset for audiovisual person search. In comparison to available audiovisual search datasets, it contains two orders of magnitude more identities at 1.9K. Additionally, the 30K manually curated face tracks contain a much larger diversity in audiovisual patterns than similar datasets, as its original movie set is composed of far more diverse videos across multiple genres and including diverse demographics. Finally, the 3.1 Million individual instances allow for modern machine learning methods techniques to be used on the APES dataset.
    }
     \label{tab:dataset-comparison}
\end{table*}

\vspace{-0.25cm}
\paragraph{Video Person Retrieval Datasets.} After many milestones achieved on image-based face retrieval, the computer vision community shifted attention into video use cases. There are many datasets and tasks related to person and face retrieval in videos. Three popular tasks have been established, including person re-identification, speaker recognition, and recently person search. Table \ref{tab:dataset-comparison} summarizes datasets for these tasks and compares them with the APES dataset. 

The first group includes datasets designed for person re-identification \cite{market, mars, psd}. These datasets usually contain many identities; however, most are composed only of cropped tracks without any visual context or audio information. Moreover, person re-identification datasets focus on surveillance scenarios, where the task is to find a target subject across an array of cameras placed in a limited area.

The second group includes speaker recognition datasets \cite{voxceleb2, voxceleb}. Datasets such as VoxCeleb2 \cite{voxceleb2} have pushed the scale up to 150K face clips. A drawback of this group of datasets is that clips are only a few seconds long and tend to contain a single face. The third group includes datasets for person retrieval. CSM \cite{csm}, for instance,  introduces the task of cast search, which consists of retrieving all the tracklets that match the identity of a portrait query. iQIYI-Video \cite{iqiyi} scales the total number of tracklets, clips, and identities. Both datasets provide a step towards visual-based person retrieval but exhibit limitations for multimodal (faces and voices) modeling. On the one hand, CSM does not provide audio streams or video context; on the other hand, iQIYI-Video contains short clips and does not associate voices with person identities. Our APES dataset mitigates these limitations by annotating long videos with people's faces, voices, and their corresponding identities.

The third group is the closest to our setup, it comprises datasets for audiovisual person search. While the Big Bang Theory datatset \cite{bigban} allows to study the same tasks as APES, it is limited to only 8 identities (which are observed mostly indoors). Additionally speech events are approximately localized using the show's transcripts. APES contains dense manual annotations for speech events and identity pairs. Sherlock \cite{sherlock} also allows for audiovisual identity search but contains only 33 identities, and its cast is composed of mostly white European adults. The Sherlock dataset also discards short segments of speech (shorter than 2 secs), this is a key limitation as our analysis shows that these short segments constitute a big portion of utterances in natural conversations. Finally, Buffy \cite{everingham2006hello} is also very small in terms of number of identities and its data lacks diversity as it was collected from only two episodes of the series.

\vspace{-0.25cm}
\paragraph{Audiovisual Learning for Identity Retrieval.} Audiovisual learning has been widely explored in the realm of multiple video tasks \cite{owens2018audio, chung2016out, ava-active-speakers, afouras2020self}, but only a few have focused their efforts on learning embeddings for person identity retrieval \cite{nagrani2018seeing, learnablepins}. Nagrani \etal \cite{nagrani2018seeing} have proposed a cross-modal strategy that can 'see voices' and 'hear faces'. It does so by learning to match a voice to one out of multiple candidate faces using a multi-way classification loss. More recently, the work in \cite{learnablepins} introduces a cross-modal approach that leverages synchronization between faces and speech. This approach assumes there is one-to-one correspondence in the audiovisual signals to form queries, positive, and negative sets and train a model via contrastive learning \cite{contrastive}. Although the method proposed in \cite{learnablepins} does not require manually annotated data, it assumes all face crops contain a person talking, an assumption that often breaks for videos in the wild. Our baseline model seizes inspiration from the success of these previous approaches in cross-modal and audiovisual learning. It leverages the newly annotated APES dataset, and its design includes a two-stream audiovisual network that jointly learns audiovisual characteristics of individuals.

\begin{figure*}[ht!]
    \begin{center}
        \includegraphics[width=0.99\textwidth]{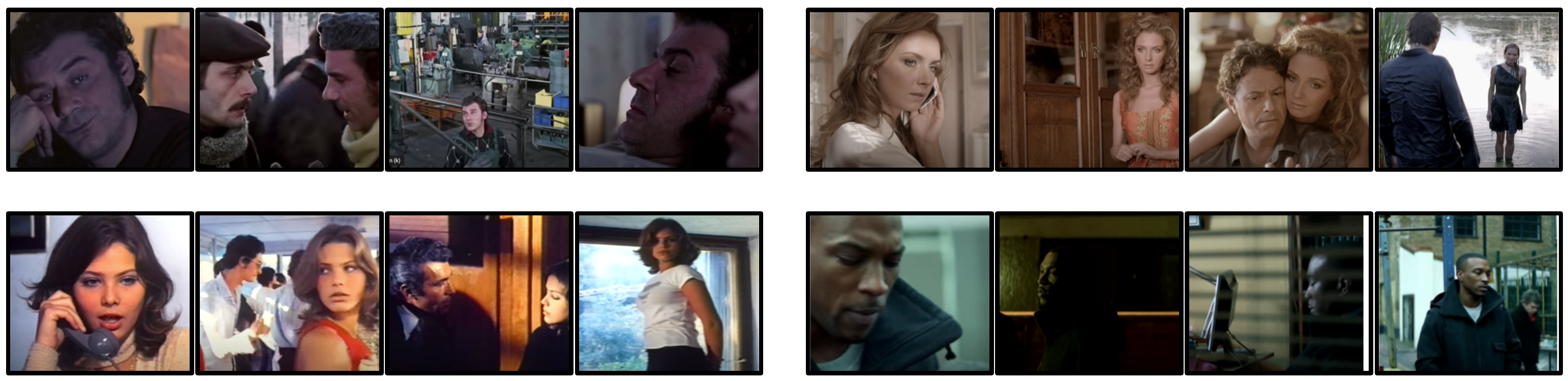}
    \end{center}
    \caption{\textbf{The APES dataset.} We illustrate a few examples from our novel APES dataset. As it can be seen, it casts many challenges for automatic audiovisual person identification and search. For instance, the appearance of each identity changes significantly across different scenes in a movie. Similarly, APES poses challenges to identify voices across time, as the environment and background sounds are always changing. In many cases, even detecting the persons' faces can be challenging due to illumination and partial occlusions.}
    \label{fig:apes-samples}
\end{figure*}

\section{APES: Audiovisual PErson Search Dataset}
\label{sec:dataset}
This section introduces the Audiovisual PErson Search (APES) dataset, which aims at fostering the study of multimodal person retrieval in videos. This new dataset consists of more than $36$ hours of untrimmed videos annotated with $1913$ unique identities. APES' videos pose many challenges, including small faces, unconventional viewpoints, as well as short segments of speech intermixed with environmental sound and soundtracks. Figure \ref{fig:apes-samples} shows a few APES samples. Here, we describe our data collection procedure and statistics of APES.

\subsection{Data Collection}

\paragraph{Video source.} We aim for a collection of videos showing faces and voices in unconstrained scenarios. While there has been a surge of video datasets in the last few years \cite{kinetics, momentsintime, charades, epickitchens}, most of them focus on action recognition on trimmed videos. As a result, it is hard to find a large video corpus with multiple instances where individuals are seen on-screen and speaking. This trend limits the availability of relevant data for audiovisual person search.

Instead of gathering user-generated videos, the AVA dataset [11] is made from movies. AVA list of movies includes productions filmed around the world in different languages and across multiple genres. It contains a diverse and extensive number of dialogue and action scenes. Another appealing property is that it provides a download mechanism \footnote{Researchers can download the full set of AVA videos at: \href{https://github.com/cvdfoundation/ava-dataset}{https://github.com/cvdfoundation/ava-dataset}} to gather the videos in the dataset; this is crucial for reproducibility and promote future research. Finally, the AVA dataset has been augmented with Active Speaker annotations \cite{ava-active-speakers}. This new set contains face tracks annotated at 20 frames per second; it also includes annotations that link speech activity with those face tracks. Consequently, we choose videos from AVA to construct the APES dataset. Our task is then to label the available face tracks and speech segments with actors' identities.

\vspace{-0.25cm}
\paragraph{Labeling face identities.}
\begin{figure*}[ht!]
    \begin{center}
        \includegraphics[width=0.99\textwidth]{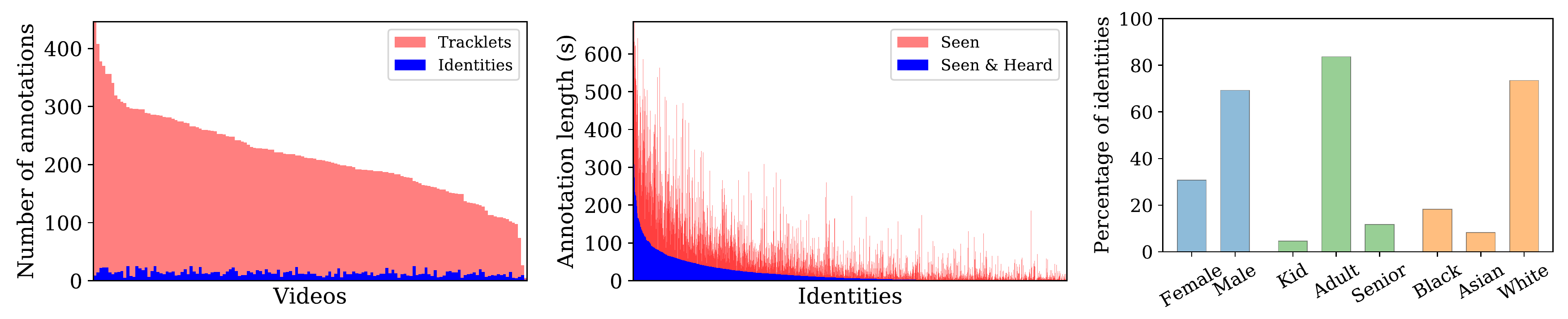}
    \end{center}
    \caption{\textbf{APES global statistics.} \textit{Left:} Distribution of number of tracklets and identities per video, the length of the tracklets roughly follows a long tail distribution, but there is not much difference in the number of identities per video. \textit{Center:} Distribution of annotation length of only seen identities, and seen \& heard identities, it is important to note that many actors are seen but remain silent, and the speaking characters get most screen time. \textit{Right:} Demographics of labeled identities in APES, sorter by gender, age group, and race. Despite representing demographics better than previous datasets \cite{bigban,iqiyi}, we still observe room for improvement towards a more balanced distribution. Nevertheless, for most demographics, the dataset contains at least 1.5K tracks and 155K face crops, a representative set for training deep learning models.}
    \label{fig:global-stats}
\end{figure*}

\begin{figure*}[h!]
    \begin{center}
        \includegraphics[width=0.99\textwidth]{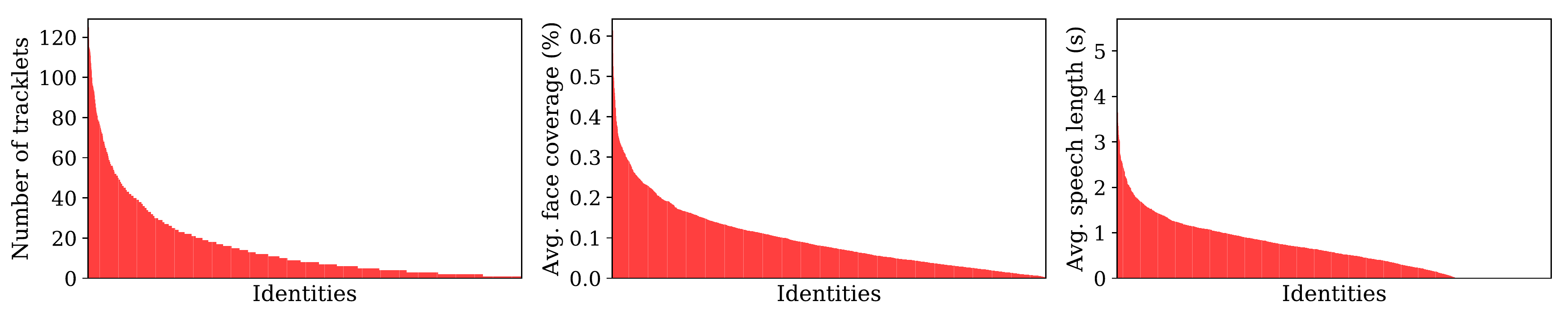}
    \end{center}
    \caption{\textbf{APES identity statistics.} We analysis 3 relevant statistics for the label distribution in APES, all of them follow a long tailed distributions \textit{Left:} Number of face tracks per identity. This distribution indicates that main characters tend to appear more often than, for instance, extras or background actors. \textit{Center:} Average face coverage per identity, computed as the mean of the relative area covered by all identity face tracks. This statistic hints that a big portion of identites are framed within close-ups, but still many identities are portrait in the background of the scene. \textit{Right:} Average length of continuous speech per identity. Interestingly, most speech segments last at most four seconds, and about 25$\%$ of the identities do not speak at all.}
    \label{fig:identity-stats}
\end{figure*}
We first downloaded a total of $144$ videos from the AVA dataset, gathered all face tracks available, and \textit{annotated} identities for a total of $33739$ face tracks. We did so in two stages. In the first stage, we addressed the identity assignment tasks per video and asked human annotators to cluster the face tracks into matching identities. To complete this first task, we employed three human annotators for $40$ hours each. We noticed two common errors during this stage: (i) false positives emerging from small faces or noisy face tracks; and (ii) false negatives that assign the same person into more than one identity cluster. We alleviated these errors by implementing a second stage to review all instances in the clustered identities and merged wrongfully split clusters. This process was a relatively shorter verification task, which annotators completed in \textit{eight} hours. At the end of this annotation stage, about $8.4\%$ of the face tracks were labeled as ambiguous; therefore, we obtained a total of $30880$ face tracks annotated among $1913$ identities.

\paragraph{Labeling voice identities.}
After labeling all face tracks with their corresponding identities, we now need to find their voice pairs. Our goal then is to cluster all voices in the videos and match their corresponding faces' original identity. In other words, we want the same person's faces and voices to share the same identity. Towards this goal, we leveraged the original annotations from the AVA-ActiveSpeakers dataset \cite{ava-active-speakers}, which contain temporal windows of speech activity associated with a face track. We mapped speech segments to their corresponding face track and assigned a common identity. We annotated $26879$ voice segments accounting for a total of $11.1$ hours of speech among the $1913$ identities.

\subsection{Dataset Statistics}
We annotated over $36$ hours from $144$ videos, including $30880$ face tracks,  $3.1M$ face bounding boxes, and $26879$ voice segments. Our labeling framework annotated $1913$ identities and discarded $2859$ ambiguous face tracks. We discuss in detail the statistics of the dataset below.

\paragraph{Global statistics.} In Figure \ref{fig:global-stats} (\textit{Left}), we observe the distribution of the number of tracklets and identities per video. The number of tracklets per video follows a long-tail distribution, and there is no correlation between the number of tracklets with the number of identities. This fact indicates that certain identities have longer coverage than others. Figure \ref{fig:global-stats} (\textit{Center}) shows the average length an identity is Seen or Seen \& Heard. Interestingly, the Seen \& Heard distribution exhibits a long-tailed distribution, with many identities being heard only very few times. Also, some identities are seen many times without speaking at all. Finally, we investigate the demographics of the dataset in Figure \ref{fig:global-stats} (\textit{Right}). To do this, we manually annotate the identities with gender, age, and race attributes. Although work needs to be done to balance samples across demographics, the survey shows that our video source has representative samples to cover the various demographic groups. For instance, for the most under-represented group, kids, APES contains more than 1.5K tracks. Moreover, APES provides significant progress from previous datasets that contain a single demographic group, \eg, iQIYI-VID \cite{iqiyi} contains only Asian celebrities, and the Big Bang Theory dataset \cite{bigban} comprises a cast limited to a single TV series. 

\paragraph{Identity statistics.} We analyze here characteristics of the annotated identities. First, in Figure \ref{fig:identity-stats} (\textit{Left}), we show the distribution of the number of tracklets per identity. We observe a long-tailed distribution where some identities, likely the main characters, have ample screen time, while others, \eg supporting cast, appear just a few times. Figure \ref{fig:identity-stats}(\textit{Center}) shows the average face coverage per identity, where we observe also a long-tail distribution. On the one hand, identities with large average face coverage include actors favored with close-ups; on the other hand, identities with low average face coverage include actors framed within a wide shot. Finally, we plot the average length of continuous speech per identity (Figure \ref{fig:identity-stats} (\textit{Right})). Naturally, different characters have different speech rhythms, and therefore the dataset exhibits a non-uniform distribution. Interestingly a big mass centers around one second of speech. This characteristic might be due to the natural dynamics of engaging dialogues. Moreover, we observe than about 25 $\%$ of the identities do not speak at all.

\section{Experimental Analysis}
\label{sec:study}

We now outline the standard evaluation setup and metrics for the APES dataset along with a baseline method that relies on a two-stream architecture for multi-modal metric learning. The first stream receives cropped faces while the second works over audio clips. Initially, each stream is optimized via triplet loss to minimize the distance between matching identities in the feature space. As highlighted in other works \cite{facenet, nagrani2018seeing, parkhi2015deep}, it is essential to acquire a clean and extensive set of triplets to achieve good performance. Below, we detail each modality of our baseline model and different subsets for training.

\vspace{-0.25cm}
\paragraph{Facial Matching.} To optimize the face matching network, we remove the classification layer from a standard Resnet-18 encoder \cite{resnet} pre-trained on ImageNet \cite{imagenet}, and fine-tune it using a triplet loss \cite{facenet}. We choose the ADAM optimizer \cite{adam} with an initial learning rate of $3\times 10^{-4}$ and learning rate annealing of $0.1$ every 30 epochs for a total of 70 epochs. We resize face crops to $124\times124$ and perform random flipping and corner cropping during training. 

\begin{figure}[t]
    \begin{center}
        \includegraphics[width=0.5\textwidth]{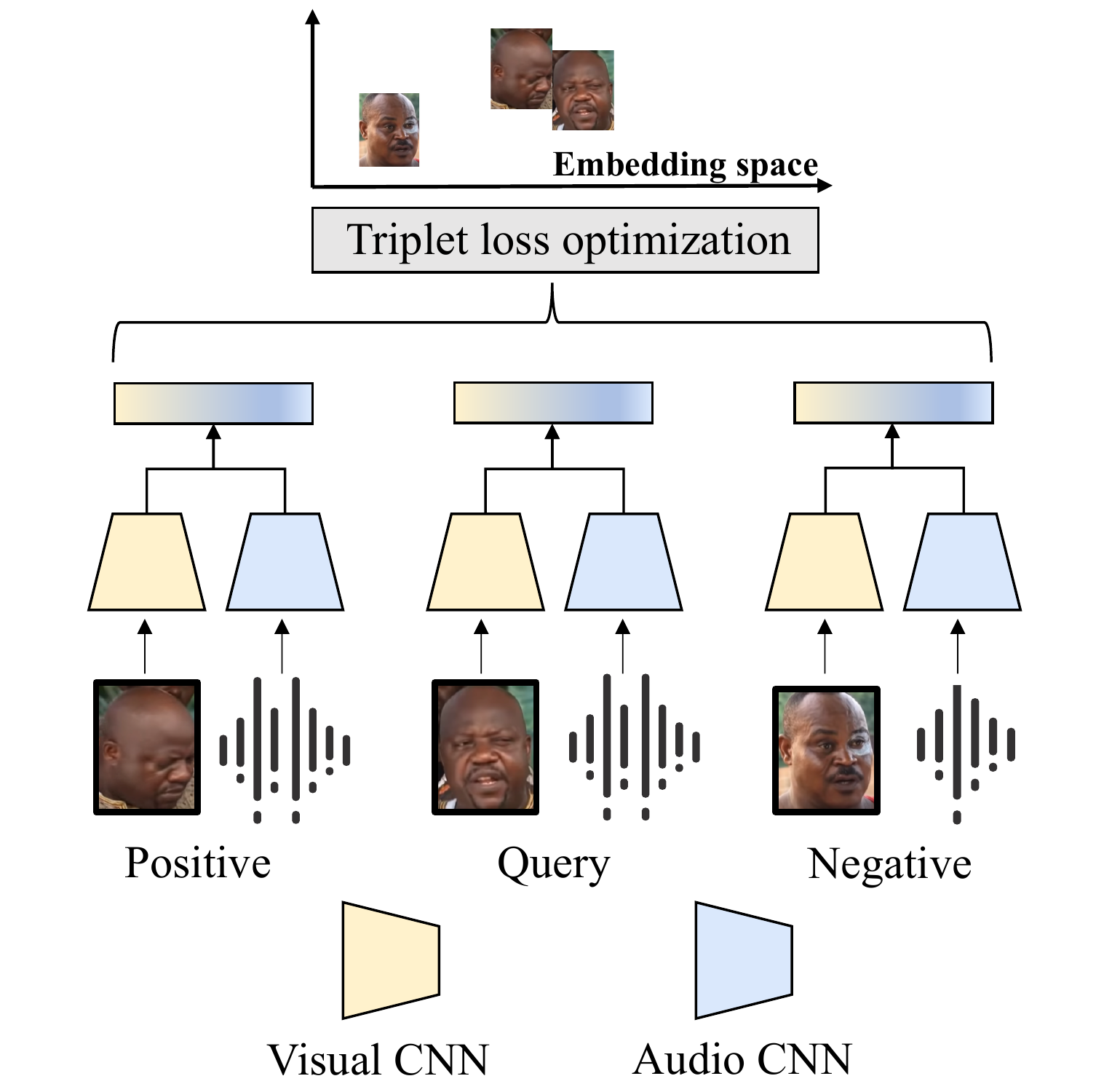}
    \end{center}
    \caption{\textbf{Creating Audiovisual Embeddings }, We use a two stream neural network to identify corresponding identities in APES. Our approach uses independent visual (light yellow) and audio (light blue) CNNs, the joint feature set is then  optimized by means of the triplet loss to obtain an embedding where corresponding identities are close located, }
    \label{fig:method}
\end{figure}

\vspace{-0.25cm}
\paragraph{Voice Matching.} Similar to the visual stream, we use a ResNet-18 model initialized with ImageNet weights and fine-tuned via triple loss learning. We follow a setup similar to \cite{ava-active-speakers} and use a Mel-spectrogram calculated from audio snippets of $0.45$ seconds length in the audio stream. We use the same hyper-parameters configuration described for the visual matching network.

\vspace{-0.25cm}
\paragraph{Cross-modal Matching.} For the audiovisual experiments, we combine the individual configurations for voice and face matching. However, we add a third loss term which optimizes the feature representation obtained from a joint embedding of audiovisual features, which we obtain via concatenation of each stream's last layer feature map. This third loss is also optimized using the triplet loss. Figure \ref{fig:method} illustrates our cross-modal baseline.

\subsection{Experimental Setup}
\paragraph{Dataset splits and Tasks.} We follow the official train-val splits defined in the original AVA-ActiveSpeaker dataset \cite{ava-active-speakers}. As not every person is actively speaking at every moment, we define two tasks: \textit{Seen \& Heard} (a person is on-screen and talking), and only \textit{Seen} (the person is on-screen but not speaking). Each of these tasks yield a corresponding training and validation subset. The Seen task subsets have $23679$ and $7989$ tracklets for training and validation respectively. Conversely, the Seen \& Heard subset, is comprised of $11862$ tracklers for training and $3582$ for validation.

\begin{table*}[t!]
    \setlength{\tabcolsep}{4pt}
    \centering
    \begin{tabular}{ l c c c c c c}
    \toprule
    & \multicolumn{3}{c}{\textbf{Seen \& Heard}} & \multicolumn{3}{c}{\textbf{Seen}} \\
    & Avg Positive & Avg Negative &  & Avg Positive   & Avg Negative  &   \\
    Sampling strategy & Tracklets & Tracklets & Clean &  Tracklets  & Tracklets &  Clean \\
    \midrule
    Weak & 1 & 11862 & \xmark & 1 & 23679 &  \xmark   \\
    Within & 10 & 112 & \cmark & 16 & 242 & \cmark  \\
    Across & 10 & 11862 & \cmark & 16 & 23679 & \cmark  \\
    \toprule
    \end{tabular}
    \caption{
        \textbf{Identity Sampling.} APES allows for multiple sampling strategies to form triplets during training. APES-IN is the simplest scenario as it samples positive and negative identities from a single video. Such sampling generates a 10:112 positive-to-negative ratio, for instance, in the case of the Seen \& Heard task. APES-Across creates a more challenging scenario, where negatives are extracted from all videos across the dataset; it is much more imbalanced problem with a 10:11862 positive-to-negative ratio. APES-weak considers an extreme scenario where a single tracklet is used to retrieve identities from the whole video collection; this setting is not only extremely unbalance 1:11862 but also challenging as the audiovisual identity from a single tracklet tend to be highly similar in appearance and sound characteristics. 
    }
     \label{tab:triple-sets}
\end{table*}

\paragraph{Identity Sampling.} The APES dataset allows us to sample positive and negative samples during training in three different ways. The most direct sampling would gather every tracklet from a single movie. In such a scenario, we will create a positive bag with the tracklets that belong to a given identity, while negative samples would be obtained from every other tracklet in the same movie.  We name this setup \textbf{Within}, a simplified configuration where we have a 1:15 ratio of positive tracklets (same identity) to negative tracklets (different identity).

While the Within modality allows us to explore the problem, it might be an overly simplified scenario. Hence, we also devise the \textbf{Across} setup, where we sample negatives identities over the full video collection, \ie, across different movies. This sampling strategy significantly changes the ratio of positive to negative tracklets to 1:150 and better resembles the natural imbalance of positive/negative identities in real-world data.

Finally, we create an extreme setup that resembles few-shot learning scenarios for identity retrieval learning. In this case, a single (or very few) positive samples are available to train our embeddings. These positive samples are sampled from the same tracklet as the query, and instances from every other tracklet in the datasets form the negative bag. We name this setup as \textbf{Weak}, which results in a strongly imbalanced subset with about 1:1500 ratio of positive to negative tracklets. A summary of these three sampling sets is presented in Table \ref{tab:triple-sets}.

\paragraph{Evaluation metrics.} Three evaluation metrics assess methods' effectiveness in APES: 

\begin{itemize}
    \item Precision at K (P@K): we estimate the precision from the top K retrieved identities for every tracklet in a video. As there are no shared identities over videos, we simply estimate the precision at K for every video, and then average for the full validation set.
   
    \item Recall at K (R@K): we estimate the recall from the top K retrieved identities for every tracklet in a video. Again, we estimate the recall at K for every video and compute the average over the full validation set.
    
    \item Mean average precision (mAP): as final and main evaluation metric, we use the mean average precision. Like in the recall and precision cases, we compute the mAP for every tracklet in a video, and the average the results for videos in the validation set.
\end{itemize}

\subsection{Benchmark Results}

\paragraph{Seen \& Heard benchmark results.}
Table \ref{tab:benchmark-results} summarizes our benchmark results for the three main configurations: Facial Matching, where we operate exclusively on visual data; Voice Matching, where we exclusively model audio data; and Cross-modal Matching, where we train and validate over the audio and visual modalities.  Overall results obtained with facial and cross-modal matching are far from perfect. Our best baseline model obtains a max mAP of 64.8\%. This result indicates that the standard triplet loss is just an initial baseline for the APES dataset and that there is ample room for improvement. The relatively high precisions at K=1 and K=5 in almost every setting, suggests the existence of a few easy positives matches for every query in the dataset. However, the relative low precision and recall at K=10 suggest that the method quickly exhibits wrong estimations as we progress through the ranking. This drawback is worst in the APES-Weak training setting, as its selection bias induces a much lower variability on the positive samples. The analysis of the recall scores highlights the importance of the additional audio information. After the network is enhanced with audio data, the recall metrics improve significantly, reaching 98.5\% at K=100. While this improvement comes at the cost of some precision, overall, the mAP shows an improvement of 0.7\%. Finally, there is only a slight difference between the Within and Across sampling settings, the former having a marginally better recall. This suggests that the massive imbalance induced in the across setting does not significantly improve the diversity of the data observed at training time and that more sophisticated sampling strategies such as hard negative mining might be required to take advantage of that extra information.

\begin{table}[t!]
    \setlength{\tabcolsep}{4pt}
     \footnotesize

    \begin{tabular}{ l | c c c | c c c | c }
    \toprule
    & R@10 & R@50 & R@100 & P@1 & P@5 & P@10 & mAP(\%) \\
    \midrule
    \textit{Random} & 12.0& 52.9 & 87.3 & 18.1 & 18.7 &  18.4 &  19.1 \\
    \midrule
    \multicolumn{8}{l}{\textbf{Facial Matching}}\\
    Weak & 38.2 & 74.5 & 94.2 & 80 & 62.3 &  49.9 &  45.4  \\
    Within & 48.1 & 86.5 & 98.1 & 87.5 & \textbf{74.0} & \textbf{62.8} & 64.1\\
    Across & \textbf{48.2} & 86.3 & 98.2 & \textbf{88.0} & 73.6 & 62.6 & 63.4 \\
    \midrule
    \multicolumn{8}{l}{\textbf{Voice Matching}}\\
    Weak & 17.4 & 59.1 & 89.4 & 29.2 & 26.9  & 23.2& 20.5  \\
    Within & 18.2 & 59.7 & 89.9 & 30.0 & 26.7 & 24.8 & 21.8  \\
    Across & 17.7 & 58.9 & 89.5 & 29.3 & 26.6 & 24.5 & 22.0 \\
    \midrule
    \multicolumn{8}{l}{\textbf{Cross-modal Matching}}\\
    Weak & 38.2 & 74.2 & 94.3 & 80.1  &  62.0 & 49.4 & 47.9  \\
    Within & 47.9 & \textbf{87.1} & \textbf{98.5} & 85.9 & 72.4 & 61.2 & \textbf{64.8}\\
    Across & 48.1 & 86.8 & 98.3 & 86.7 & 73.3 &  62.2 & 63.6  \\    
    \bottomrule
 
    \end{tabular}

    \caption{
        \textbf{Benchmark results for Seen \& Heard task.} We measure recall at K (R$@$K), precision at K (P$@$K), and the mean Averge Precision (mAP).
        We observe that the Cross-modal matching baseline slightly outperforms the facial matching model (by $0.7\%$ mAP). This result suggests that the audio signal is helpful to the re-identification process, especially increasing the recall of the proposed method. The relative low precision for larger K's suggest that correspondences are easy to build for a few images, but much harder as candidates become more diverse. We also observe that identifying persons using only voice is a much harder task. 
    }
    \label{tab:benchmark-results}
\end{table}

\begin{table}[t!]
    \setlength{\tabcolsep}{4pt}
     \footnotesize

    \begin{tabular}{ l | c c c | c c c | c }
    \toprule
    & \multicolumn{7}{c}{\textbf{Seen Full Set}} \\
    & R@10 & R@50 & R@100 & P@1 & P@5 & P@10 & mAP(\%) \\
    \midrule
    \textit{Random} & 5.1 & 26.0 & 52.7 & 17.2 & 17.1 & 16.9 & 17.1 \\
    \midrule
    \multicolumn{8}{l}{\textbf{Facial Matching}}\\
    Weak & 26.9 & 54.5 & 74.5 & 82.2 & 66.0 & 54.6 &  39.3  \\
    Within & 33.1 & 70.5 & 87.4& 87.8 & 77.3 & 68.7 & 58.8\\
    Across & 33.8 & 70.4 &87.6 & 88.2 & 77.6 & 68.6 & 59.4  \\
    \bottomrule
    \end{tabular}

    \caption{
        \textbf{Benchmark for the Seen Set.} We use the same metrics as in the Seen \& Heard setting. We find that this configuration is slightly harder despite having more data. This counter-intuitive result can be explained as movies often rely on shots where a active-speakers are portray in close-ups. In short, the Seen task has more data available for training, but also more challenging scenarios.
    }
    \label{tab:benchmark-results-seen}
\end{table}
\begin{figure*}[t!]
    \begin{center}
        \includegraphics[width=0.98\textwidth]{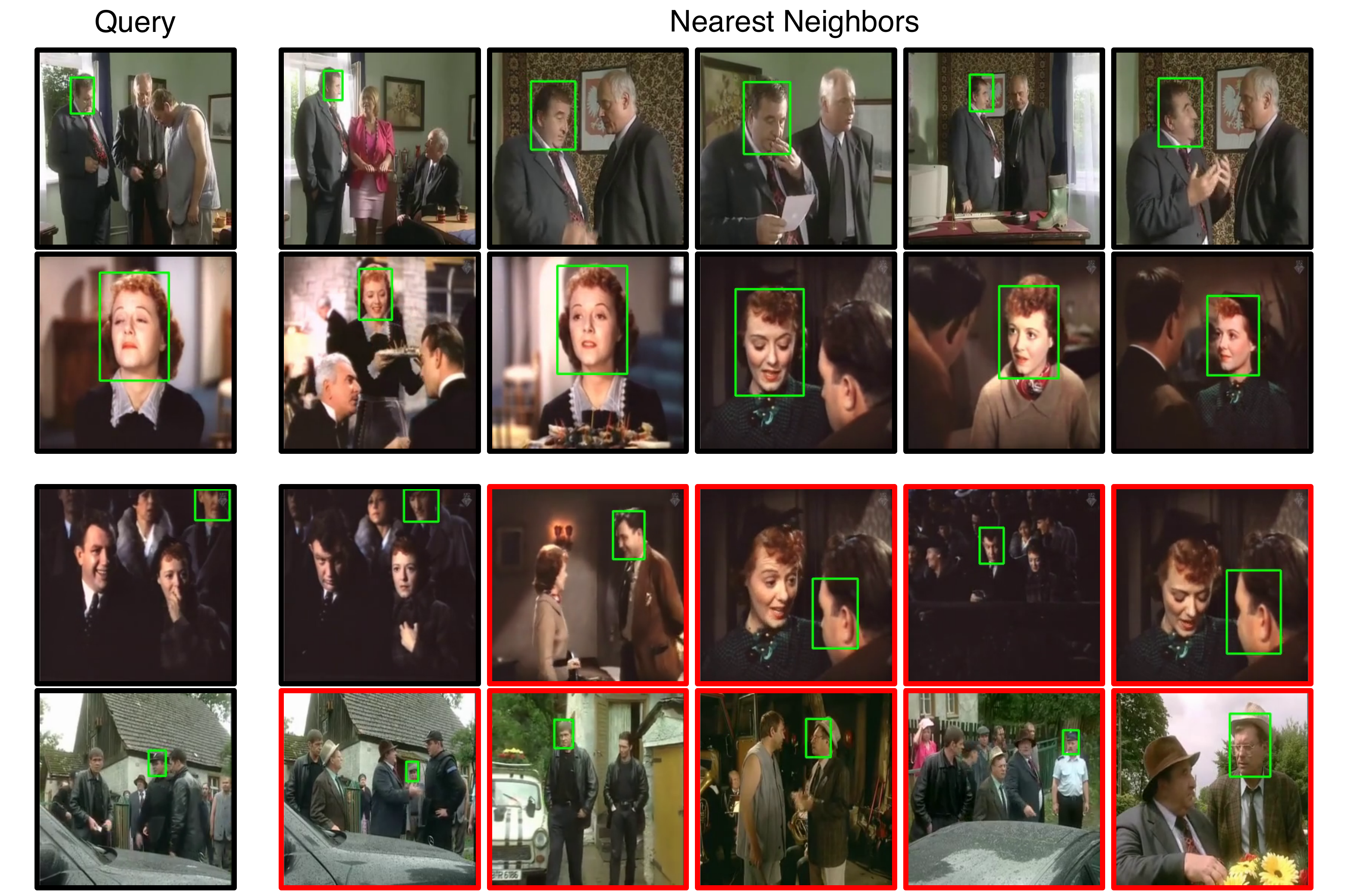}
    \end{center}
    \caption{\textbf{Qualitative results.} We showcase easy (top two rows) and hard (bottom two rows) examples in our benchmark. Green bounding boxes indicate the localization of the queried and retrieved faces, and red borders around the frame indicates a false positive. We empirically observe that queries from large faces within dialogue scenes tend to be easier to retrieve. In contrast, queries from small faces and cases where the subject is moving are challenging to our baseline.}
    \label{fig:qualitative-results}
\end{figure*}

\paragraph{Seen benchmark results.}
We empirically found that fusing modalities with noisy audio data (original splits) provides no improvement. As outlined before, our experiments suggest that audio models are highly sensitive to noisy speech annotations and do not converge if there is large uncertainty in the ground truth. Under this circumstances the optimization just learns to ignore the audio cues, and yields the same performance as the visual-only setting.

Despite this we report our results of a facial matching for the Seen task, as it will serve as baseline for future works that can handle noise speech data. Table \ref{tab:benchmark-results-seen} contains the benchmark for the Seen task. We find that this task is actually harder than the Seen \& Heard task, despite having more available data. We explain this result as movies typically depict speakers over large portions of the screen and offer a clear view angle to him/her, which results on average larger faces with less noise in the Seen \& Heard task, and smaller more challenging faces in the Seen. In other words, we hypothesize that speaking faces are easier to re-identify as they are usually framed within close-ups. This bias makes it harder to find every matching tracklet for the identity (thus reducing recall).

\subsection{Qualitative Results}
We showcase easy and hard instances for our baseline in Figure \ref{fig:qualitative-results}. Every row shows a query in the left, and the top five nearest neighbors on the right. The first two rows, shows instances where our baseline model correctly retrieves instances of the same person. We have empirically noticed that instances where query faces are large, \eg in close-ups and medium shots from dialogue scenes, our baseline model tends to provide a very good ranking to the retrieved instances. The two rows from the bottom illustrate hard cases where the baseline model fails. We have seen that small faces, poor illumination, occlusion, and subjects in motion present a challenging scenario to our baseline.

\section{Conclusion}
We introduce APES, a new dataset for audiovisual person search. We compare APES with existing datasets for person identity analysis and show that it complements previous datasets in that those have mainly focused on visual analysis only. We include benchmarks for two tasks \textit{Seen} and \textit{Seen \& Heard} to showcase the value of curating a new audiovisual person search dataset. We believe in the crucial role of datasets at measuring progress in computer vision; therefore, we are committed to releasing APES to enable further development, research, and benchmarking in the field of audiovisual person identity understanding.

\clearpage
{\small
\bibliographystyle{ieee_fullname}
\bibliography{apes}

\begin{thebibliography}{10}\itemsep=-1pt

\bibitem{afouras2020self}
Triantafyllos Afouras, Andrew Owens, Joon~Son Chung, and Andrew Zisserman.
\newblock Self-supervised learning of audio-visual objects from video.
\newblock {\em arXiv preprint arXiv:2008.04237}, 2020.

\bibitem{bigban}
Martin Bauml, Makarand Tapaswi, and Rainer Stiefelhagen.
\newblock Semi-supervised learning with constraints for person identification
  in multimedia data.
\newblock In {\em Proceedings of the IEEE Conference on Computer Vision and
  Pattern Recognition}, pages 3602--3609, 2013.

\bibitem{bruce1986understanding}
Vicki Bruce and Andy Young.
\newblock Understanding face recognition.
\newblock {\em British journal of psychology}, 77(3):305--327, 1986.

\bibitem{voxceleb2}
Joon~Son Chung, Arsha Nagrani, and Andrew Zisserman.
\newblock Voxceleb2: Deep speaker recognition.
\newblock {\em arXiv preprint arXiv:1806.05622}, 2018.

\bibitem{chung2016out}
Joon~Son Chung and Andrew Zisserman.
\newblock Out of time: automated lip sync in the wild.
\newblock In {\em ACCV}, 2016.

\bibitem{epickitchens}
Dima Damen, Hazel Doughty, Giovanni Maria~Farinella, Sanja Fidler, Antonino
  Furnari, Evangelos Kazakos, Davide Moltisanti, Jonathan Munro, Toby Perrett,
  Will Price, et~al.
\newblock Scaling egocentric vision: The epic-kitchens dataset.
\newblock In {\em Proceedings of the European Conference on Computer Vision
  (ECCV)}, pages 720--736, 2018.

\bibitem{imagenet}
Jia Deng, Wei Dong, Richard Socher, Li-Jia Li, Kai Li, and Li Fei-Fei.
\newblock Imagenet: A large-scale hierarchical image database.
\newblock In {\em 2009 IEEE conference on computer vision and pattern
  recognition}, pages 248--255. Ieee, 2009.

\bibitem{everingham2006hello}
Mark Everingham, Josef Sivic, and Andrew Zisserman.
\newblock Hello! my name is... buffy''--automatic naming of characters in tv
  video.
\newblock In {\em BMVC}, volume~2, page~6, 2006.

\bibitem{farenzena2010person}
Michela Farenzena, Loris Bazzani, Alessandro Perina, Vittorio Murino, and Marco
  Cristani.
\newblock Person re-identification by symmetry-driven accumulation of local
  features.
\newblock In {\em 2010 IEEE Computer Society Conference on Computer Vision and
  Pattern Recognition}, pages 2360--2367. IEEE, 2010.

\bibitem{gheissari2006person}
Niloofar Gheissari, Thomas~B Sebastian, and Richard Hartley.
\newblock Person reidentification using spatiotemporal appearance.
\newblock In {\em 2006 IEEE Computer Society Conference on Computer Vision and
  Pattern Recognition (CVPR'06)}, volume~2, pages 1528--1535. IEEE, 2006.

\bibitem{grother2017face}
Patrick~J Grother, Mei~L Ngan, and George~W Quinn.
\newblock Face in video evaluation (five) face recognition of non-cooperative
  subjects.
\newblock Technical report, 2017.

\bibitem{contrastive}
Raia Hadsell, Sumit Chopra, and Yann LeCun.
\newblock Dimensionality reduction by learning an invariant mapping.
\newblock In {\em 2006 IEEE Computer Society Conference on Computer Vision and
  Pattern Recognition (CVPR'06)}, volume~2, pages 1735--1742. IEEE, 2006.

\bibitem{resnet}
Kaiming He, Xiangyu Zhang, Shaoqing Ren, and Jian Sun.
\newblock Deep residual learning for image recognition.
\newblock In {\em Proceedings of the IEEE conference on computer vision and
  pattern recognition}, pages 770--778, 2016.

\bibitem{csm}
Qingqiu Huang, Wentao Liu, and Dahua Lin.
\newblock Person search in videos with one portrait through visual and temporal
  links.
\newblock In {\em Proceedings of the European Conference on Computer Vision
  (ECCV)}, pages 425--441, 2018.

\bibitem{kinetics}
Will Kay, Joao Carreira, Karen Simonyan, Brian Zhang, Chloe Hillier, Sudheendra
  Vijayanarasimhan, Fabio Viola, Tim Green, Trevor Back, Paul Natsev, et~al.
\newblock The kinetics human action video dataset.
\newblock {\em arXiv preprint arXiv:1705.06950}, 2017.

\bibitem{adam}
Diederik~P Kingma and Jimmy Ba.
\newblock Adam: A method for stochastic optimization.
\newblock {\em arXiv preprint arXiv:1412.6980}, 2014.

\bibitem{iqiyi}
Yuanliu Liu, Bo Peng, Peipei Shi, He Yan, Yong Zhou, Bing Han, Yi Zheng, Chao
  Lin, Jianbin Jiang, Yin Fan, et~al.
\newblock iqiyi-vid: A large dataset for multi-modal person identification.
\newblock {\em arXiv preprint arXiv:1811.07548}, 2018.

\bibitem{momentsintime}
Mathew Monfort, Alex Andonian, Bolei Zhou, Kandan Ramakrishnan, Sarah~Adel
  Bargal, Tom Yan, Lisa Brown, Quanfu Fan, Dan Gutfruend, Carl Vondrick, et~al.
\newblock Moments in time dataset: one million videos for event understanding.
\newblock {\em arXiv preprint arXiv:1801.03150}, 2018.

\bibitem{learnablepins}
Arsha Nagrani, Samuel Albanie, and Andrew Zisserman.
\newblock Learnable pins: Cross-modal embeddings for person identity.
\newblock In {\em Proceedings of the European Conference on Computer Vision
  (ECCV)}, pages 71--88, 2018.

\bibitem{nagrani2018seeing}
Arsha Nagrani, Samuel Albanie, and Andrew Zisserman.
\newblock Seeing voices and hearing faces: Cross-modal biometric matching.
\newblock In {\em Proceedings of the IEEE conference on computer vision and
  pattern recognition}, pages 8427--8436, 2018.

\bibitem{voxceleb}
Arsha Nagrani, Joon~Son Chung, and Andrew Zisserman.
\newblock Voxceleb: a large-scale speaker identification dataset.
\newblock {\em arXiv preprint arXiv:1706.08612}, 2017.

\bibitem{sherlock}
Arsha Nagrani and Andrew Zisserman.
\newblock From benedict cumberbatch to sherlock holmes: Character
  identification in tv series without a script.
\newblock {\em arXiv preprint arXiv:1801.10442}, 2018.

\bibitem{owens2018audio}
Andrew Owens and Alexei~A Efros.
\newblock Audio-visual scene analysis with self-supervised multisensory
  features.
\newblock In {\em ECCV}, 2018.

\bibitem{parkhi2015deep}
Omkar~M Parkhi, Andrea Vedaldi, and Andrew Zisserman.
\newblock Deep face recognition.
\newblock 2015.

\bibitem{ava-active-speakers}
Joseph Roth, Sourish Chaudhuri, Ondrej Klejch, Radhika Marvin, Andrew
  Gallagher, Liat Kaver, Sharadh Ramaswamy, Arkadiusz Stopczynski, Cordelia
  Schmid, Zhonghua Xi, et~al.
\newblock Ava active speaker: An audio-visual dataset for active speaker
  detection.
\newblock In {\em ICASSP 2020-2020 IEEE International Conference on Acoustics,
  Speech and Signal Processing (ICASSP)}, pages 4492--4496. IEEE, 2020.

\bibitem{facenet}
Florian Schroff, Dmitry Kalenichenko, and James Philbin.
\newblock Facenet: A unified embedding for face recognition and clustering.
\newblock In {\em Proceedings of the IEEE conference on computer vision and
  pattern recognition}, pages 815--823, 2015.

\bibitem{charades}
Gunnar~A Sigurdsson, G{\"u}l Varol, Xiaolong Wang, Ali Farhadi, Ivan Laptev,
  and Abhinav Gupta.
\newblock Hollywood in homes: Crowdsourcing data collection for activity
  understanding.
\newblock In {\em European Conference on Computer Vision}, pages 510--526.
  Springer, 2016.

\bibitem{tapaswi2019video}
Makarand Tapaswi, Marc~T Law, and Sanja Fidler.
\newblock Video face clustering with unknown number of clusters.
\newblock In {\em Proceedings of the IEEE International Conference on Computer
  Vision}, pages 5027--5036, 2019.

\bibitem{psd}
Tong Xiao, Shuang Li, Bochao Wang, Liang Lin, and Xiaogang Wang.
\newblock Joint detection and identification feature learning for person
  search.
\newblock In {\em Proceedings of the IEEE Conference on Computer Vision and
  Pattern Recognition}, pages 3415--3424, 2017.

\bibitem{zhang2019fully}
Aonan Zhang, Quan Wang, Zhenyao Zhu, John Paisley, and Chong Wang.
\newblock Fully supervised speaker diarization.
\newblock In {\em ICASSP 2019-2019 IEEE International Conference on Acoustics,
  Speech and Signal Processing (ICASSP)}, pages 6301--6305. IEEE, 2019.

\bibitem{mars}
Liang Zheng, Zhi Bie, Yifan Sun, Jingdong Wang, Chi Su, Shengjin Wang, and Qi
  Tian.
\newblock Mars: A video benchmark for large-scale person re-identification.
\newblock In {\em European Conference on Computer Vision}, pages 868--884.
  Springer, 2016.

\bibitem{market}
Liang Zheng, Liyue Shen, Lu Tian, Shengjin Wang, Jingdong Wang, and Qi Tian.
\newblock Scalable person re-identification: A benchmark.
\newblock In {\em Proceedings of the IEEE international conference on computer
  vision}, pages 1116--1124, 2015.

\end{thebibliography}
}

\end{document}